\documentclass{article}
\usepackage{ijcai20}

\usepackage{times}
\usepackage{soul}
\usepackage{url}
\usepackage[hidelinks]{hyperref}
\usepackage[utf8]{inputenc}
\usepackage{caption}
\usepackage{subcaption}
\usepackage{graphicx}
\usepackage{amsmath}
\usepackage{amsthm}
\usepackage{booktabs}
\usepackage{algorithm}
\usepackage{algorithmic}
\usepackage{amssymb}
\usepackage{multirow}
\begin{document}

\title{Self-Supervised Tuning for Few-Shot Segmentation}

\author{
	Kai Zhu, Wei Zhai, Zheng-Jun Zha\footnotemark[1], Yang Cao\footnotemark[1]
	\affiliations
	University of Science and Technology of China \emails
	\{zkzy, wzhai056\}@mail.ustc.edu.cn, \{zhazj, forrest\}@ustc.edu.cn
}

\maketitle

\renewcommand{\thefootnote}{\fnsymbol{footnote}}
\footnotetext[1]{Corresponding author}

\maketitle

\begin{abstract}
	Few-shot segmentation aims at assigning a category label to each image pixel with few annotated samples. It is a 
	challenging task since the dense prediction can only be achieved under the guidance of latent features defined 
	by sparse annotations. Existing meta-learning method tends to fail in generating category-specifically 
	discriminative descriptor when the visual features extracted from support images are marginalized in embedding 
	space. To address this issue, this paper presents an adaptive tuning framework, in which the distribution of 
	latent features across different episodes is dynamically adjusted based on a self-segmentation scheme, augmenting 
	category-specific descriptors for label prediction. Specifically, a novel self-supervised inner-loop is firstly 
	devised as the base learner to extract the underlying semantic features from the support image. Then, gradient 
	maps are calculated by back-propagating self-supervised loss through the obtained features, and leveraged as 
	guidance for augmenting the corresponding elements in embedding space. Finally, with the ability to 
	continuously learn from different episodes, an optimization-based meta-learner is adopted as outer loop of our 
	proposed framework to gradually refine the segmentation results. Extensive experiments on benchmark PASCAL-$5^{i}$ 
	and COCO-$20^{i}$ datasets demonstrate the superiority of our proposed method over state-of-the-art. 
\end{abstract}

\section{Introduction}

Recently, semantic segmentation models \cite{long2015fully} have made great progress under full supervision, 
and some of them have even surpassed the level of human recognition. However, 
when the learned model is applied to a new segmentation task, it takes great cost to collect 
a large amount of full annotated data in pixel level. Furthermore, samples are not available 
in large quantities in some areas such as health care, security and so on. To address this problem, 
various few-shot segmentation methods are proposed.

One solution for solving few-shot segmentation \cite{shaban2017one} is meta-learning \cite{munkhdalai2017meta}, 
whose general idea is to utilize a large number of episodes similar to target task to learn a meta learner that 
generate an initial segmentation model, and a base learner that quickly tunes the 
model with few samples. In most methods, a powerful feature extractor with good migration ability is 
provided for the meta learner to map the query and support images into a shared embedding space. 
And the base learner generates a category-specific descriptor with the support set. The similarity between 
the query's feature maps and the descriptor is measured under a parametric or nonparametirc metric, and leveraged as 
guidance for dense prediction of the query branch.

\begin{figure}[t]  
	\centering  
	\includegraphics[width=82mm]{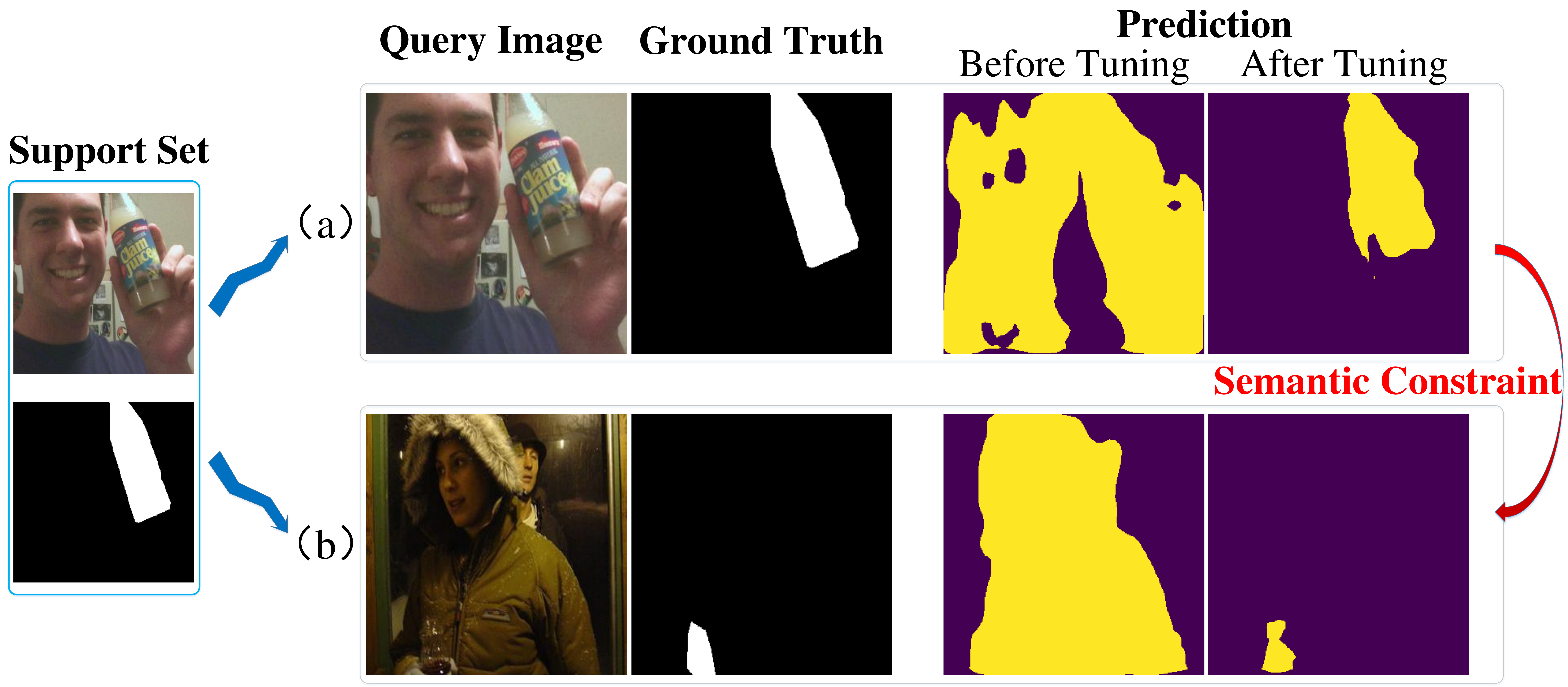}\\  
	\caption{Comparison of prediction results with and without the self-supervised tuning framework. The 
	self-supervised tuning process provides the category-specific semantic constraint to facilitate the features of 
	the person and the bottle more discriminative, thereby improving the few-shot segmentation performance 
	of corresponding categories. (a) Self-segmentation results. The support set acts as the supervision to 
	segment the support image itself. Before tuning, the pereon is incorrectly identified as the bottle even 
	in self-segmentation case. (b) Common one-shot segmentation. The query and support images are different 
	objects belonging to the same category. After tuning by the self-supervised branch, the bottle regions 
	are distinguished from the person.}  
	\label{fig:first}  
\end{figure}

However, when the low-level visual features of foreground objects extracted from the support images are too marginalized 
in the embedding space \cite{zhang2014robust}, the generated descriptor by base learner is not category-specifically discriminative. In this 
case, the regions to be segmented in the query image may be ignored or even confused with other categories in the 
background. For example, as shown in Fig. \ref{fig:first}, the bottle-specific descriptor generated from the support 
image is unable to identify the bottle itself in the same image, and therefore inapplicable for bottles in 
other complicated scenarios to be segmented. 

To address the issue, this paper presents an adaptive tuning framework for few-shot segmentation, in which 
the marginalized distributions of latent features are dynamically adjusted by a self-supervised scheme. The 
core of the proposed framework is the base learner driven by a self-segmentation task, which augments 
category-specific constraint on each new episode and facilitate successive labeling. Specifically, the base 
learner is designed as an inner-loop task, where the support images are segmented under the supervision of 
the presented masks of input support images. By back-propagating self-supervised loss through the support 
feature map, the corresponding gradients are calculated and used as guidance for augmenting each element 
in the support embedding space.

Moreover, the self-segmentation task can be considered as a special case of few-shot segmentation with the 
query and support image to be the same. Therefore, we also utilize the resulting loss of the task (we call it 
auxiliary loss in this paper) to promote the training process. Since auxiliary loss is equivalent to performing 
data enhancement to annotated samples, it can involve more information for training with the same number of 
iterations. The evaluation is performed on the prevailing public benchmark datasets PASCAL-$5^{i}$ and COCO-$20^{i}$,  
and the results demonstrate above-par segmentation performance and generalization ability of our proposed method.

Our main contributions are summarized as follows:

1. An adaptive tuning framework is proposed for few-shot segmentation, in which the marginalized distributions of latent 
category features are dynamically adjusted by a self-supervised scheme.

2. A novel base learner driven by a self-segmentation task is proposed, which augments category-specific feature description 
on each new episode, resulting in better performance on label prediction.

3. Experimental results on two public benchmark datasets PASCAL-$5^{i}$ and COCO-$20^{i}$ demonstrate the superiority 
of our proposed method over SOTA.

\begin{figure*}[t]  
	\centering  
	\includegraphics[width=146mm]{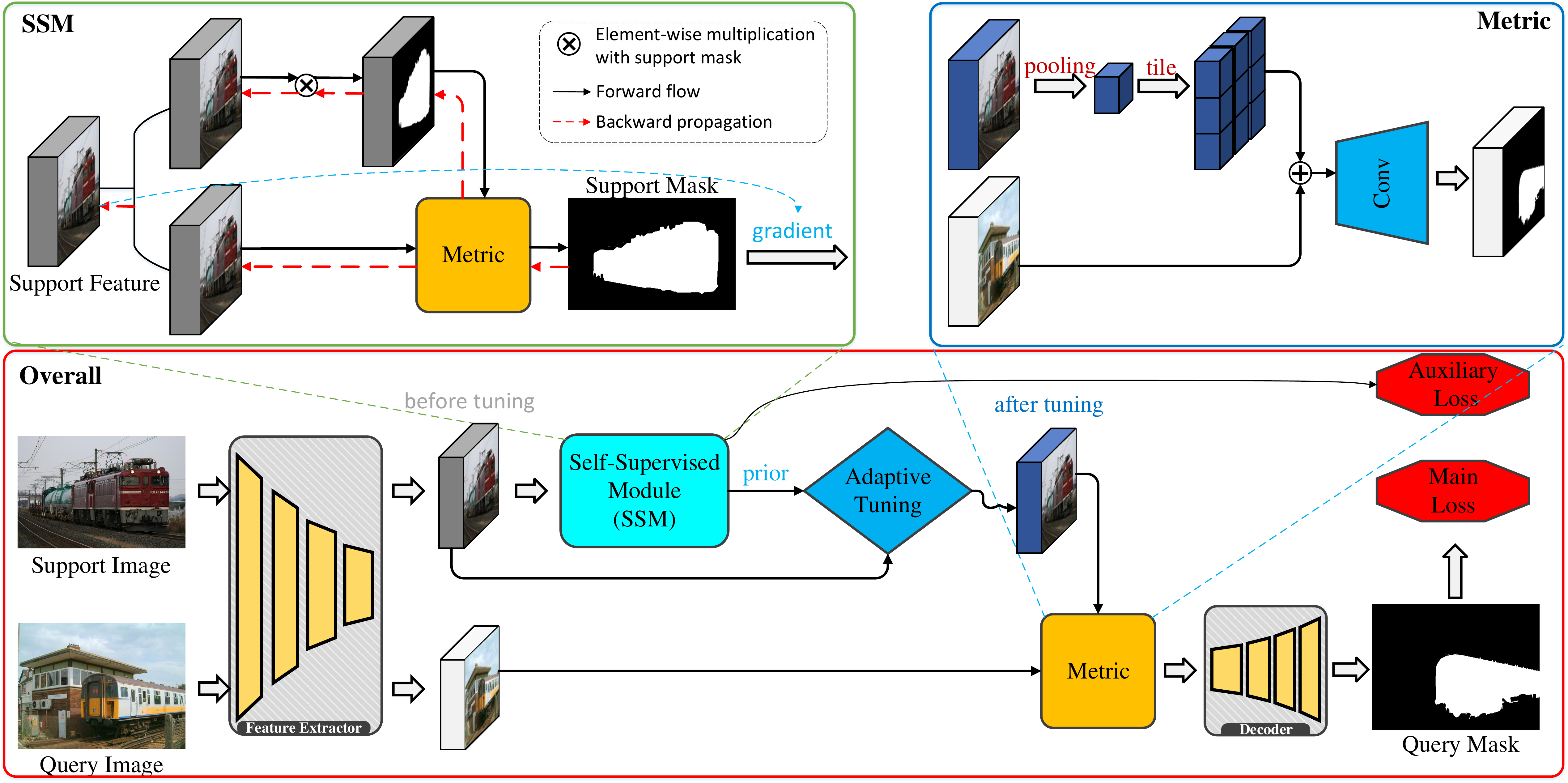}\\  
	\caption{Overall architecture of our model. It mainly consists of an adaptive
	tuning mechanism, a self-supervised base learner and a deep non-linear metric.}  
	\label{fig:model}  
\end{figure*}

\section{Related Work}

\textbf{Few-shot learning.} Few-shot learning has recently 
received a lot of attention and substantial progress has been made based on meta-learning. 
Generally, these methods can be divided into three categories. Metric-based methods focus
on the similarity metric function over the embeddings \cite{snell2017prototypical}. 
Model-based method mainly utilizes the internal architecture of the network (such as memory module 
\cite{santoro2016one}, etc.) to realize the rapid parameter adpatation in new categories. The optimization-based 
method aims at learning a update scheme for base learner \cite{munkhdalai2017meta} in each episode. 
In the latest study, \cite{lee2019meta} and \cite{bertinetto2018meta} introduce machine learning methods such as 
SVM and ridge regression into the inner loop of the base learner, and \cite{rusu2018meta} directly replaces the 
inner loop with an encoded-decode network. These methods have achieved state-of-the-art performance in few-shot 
classification task. Our model also takes inspiration of them.

\textbf{Semantic Segmentation.} Semantic segmentation is an important task in computer vision and FCNs 
\cite{long2015fully} have greatly promoted the development of the field. After that, DeepLabV3 
\cite{chen2017rethinking} and PSPNet \cite{zhao2017pyramid} propose different global contextual modules, 
which pay more attention to the scale change and global information in the segmentation process. 
\cite{zhu2019asymmetric} considers the full-image dependencies from all pixels based on Non-local 
Networks, which shows superior performance in terms of reasoning.

\textbf{Few-shot Semantic Segmentation.} While the work on few-shot learning is quite extensive, 
the research on few-shot segmentation \cite{zhang2019pyramid} \cite{hu2019attention} has 
been presented only recently. \cite{shaban2017one} 
first proposes the definition and task of one-shot segmentation. Following this, \cite{rakelly2018few} 
solves the problem with sparse pixel-wise annotations, and then extends their method to interactive 
image segmentation and video object segmentation. \cite{dong2018few} generalizes the few-shot semantic 
segmentation problem from 1-way (class) to N-way (classes). \cite{zhang2019canet} introduces an attention 
mechanism to effectively fuse information from multiple support examples and proposes an iterative 
optimization module to refine the predicted results. \cite{nguyen2019feature} and \cite{wang2019panet} 
leverage the annotations of the support images as supervision in different ways. Different from 
\cite{tian2019differentiable} which employs the base learner directly with linear 
classifier, our method devises a novel self-supervised base learner which is more intuitive and 
effective for few-shot segmentation. Compared to \cite{nguyen2019feature}, the support images are used for 
supervision in both training and test stages of our framework. 

\section{Problem Description}
\label{problem}
Here we define an input triple $Tr_{i} = (Q_{i}, S_{i}, T_{S}^{i})$, a label $T_{Q}^{i}$ and a relation function 
$F:A_{i}=F(Q_{i}, S_{i}, T_{R}^{i};\theta)$, where $Q_{i}$ and $S_{i}$ are the query and support images containing
objects of the same class i, correspondingly. $T_{S}^{i}$ and $T_{Q}^{i}$ are the pixel-wise labels corresponding 
to the $i^{th}$ class objects in $S_{i}$ and $Q_{i}$. $A_{i}^{j}$ is the actual segmentation result, and $\theta$ 
is all parameters to be optimized in function F. Our task is to randomly sample triples from the dataset, train 
and optimize $\theta$, thus minimizing the loss function $L$:
\begin{equation}
\theta _{\ast } = arg \min\limits_{\theta} L(A_{i}, T_{Q}^{i}).
\end{equation}

We expect that the relationship function $F$ can segment object regions of the same class in another target image 
each time it sees few support images belonging to a new class. This is the embodiment of the meaning of few-shot 
segmentation. It should be mentioned that the classes sampled by the test set are not present in the training set, 
that is, $U_{train} \bigcap U_{test}=\O $. The relation function $F$ in this problem is implemented by the model 
detailed in Sec. \ref{section:meta}.



\section{Method}
\subsection{Model Overview}
\label{overview}
Different from the existing methods, this paper proposes a novel self-supervised tuning framework for few shot segmentation, which is mainly composed of an adaptive tuning mechanism, a self-supervised base learner and a
meta learner. These three components will be illustrated in the next three subsections. In this subsection, we mainly present the description of the whole framework mathematically, and the symbolic representation is consistent with that in Sec. \ref{problem}.

As shown in Fig. \ref{fig:model}, a Siamese network $f_{e}$ \cite{koch2015siamese} is first proposed to extract the features of input query and support images \cite{zhang2012attribute}. The mechanism of parameters sharing not only promotes optimization but also reduces the amount of calculation. In this
step, the latent visual feature representations $R_{q}$ and $R_{s}$ are obtained as follows:
\begin{align}
	R_{q}=f_{e}(Q_{i};\theta_{e}),\\
	R_{s}=f_{e}(S_{i};\theta_{e}).
\end{align}
Here $\theta_{e}$ is the learnable parameter of the sharing encoder.
To dynamically adjust the latent features, we firstly devise a novel self-supervised inner-loop as the base learner ($f_{b}$) to exploit the underlying semantic information ($\theta_{s}$). 
\begin{equation}
	\theta_{s} = f_{b}(R_{s},T_{S}^{i};\theta_{b})
\end{equation}
Then the distribution of low-level visual features is tuned ($f_{t}$) according to the high-level category-specific cues obtained above:
\begin{equation}
	R_{s}^{\prime} = f_{t}(R_{s};\theta_{s})
\end{equation}
Inspired by Relation Network \cite{sung2018learning}, a deep non-linear metric $f_{m}$ is introduced into our meta learner.
It measures the similarity between the feature map of query image and the tuned feature, and accordingly determines regions of interest in query images. Finally, a segmentation decoder $f_{d}$ is utilized to refine the response area to the original image size:
\begin{align}
	\label{formula:compare}
	M &= f_{m}(R_{q}, R_{s}^{\prime};\theta_{m}) \\
	S &= f_{d}(M; \theta_{d})
\end{align}
Similarly, $\theta_{m}$ and $\theta_{d}$ stand for the parameters of metric and decoding part. With
the ability to continuously learn from different episodes, our meta learner gradually optimizes the whole process above (outer loop) 
and improves the performance of base learner, metric and decoder.

\subsection{Self-Supervised Tuning Scheme}
The self-supervised tuning scheme is the core of our proposed method, which is implemented by a base learner driven by self-segmentation and an adaptive tuning module. Different from the existing base learner used in \cite{lee2019meta} \cite{bertinetto2018meta}, our proposed base learner is generated under the supervision of the presented masks of input support images. This method origins from an intuitive idea, that is, the premise of identifying the regions with the same category as target objects is to identify the object itself first. Therefore, we duplicate the features of the support image as the two inputs of Eq. \ref{formula:compare} in Sec. \ref{overview} and calculate the standard cross-entropy loss ($L_{cross}$) with the corresponding support mask:
\begin{align}
	M_{sup} &= f_{m}(R_{s}, R_{s};\theta_{m}) \\
	S_{sup} &= f_{d}(M_{sup}; \theta_{d}) \\
	L_{sup} &= L_{cross}(S_{sup}, T_{S}^{i})
\end{align}

To exploit the above information to better implement the segmentation task of the query branch, the marginalized distribution is adjusted in embedding space based on the category-specific semantic constraint. Specifically, a gradient map that is calculated by back-propagating self-supervised loss through the support feature map is used as a guidance for augmenting each element in the support embedding space. Mathematically, 
\begin{equation}
	R_{s}^{\prime} = R_{s} - \frac{\partial L_{sup}}{\partial R_{s}}
\end{equation}
Note that only the feature representation is updated here and the network parameters are unchanged.

\begin{table}[t]	
	\centering
	\begin{subtable}[t]{3.3in}
		\centering
		\begin{tabular}{lcccccc}
			\toprule[1pt]
			\toprule[1pt]
			+SSM & +SS Loss & i=0 & i=1 & i=2 & i=3 & mean\\
			\midrule
			& & 49.6 & 62.6 & 48.7 & 48.0 & 52.2  \\
			\checkmark & & 53.2  & 63.6 & 48.7 & 47.9 & 53.4  \\
			& \checkmark & 51.1 & 64.9  & 51.9  & 50.2 & 54.5  \\
			\checkmark & \checkmark & \textbf{54.4}  & \textbf{66.6} & \textbf{56.2} & \textbf{52.5} & \textbf{57.4} \\
			\bottomrule[1pt]
			\bottomrule[1pt]
		\end{tabular}
		\caption{Results for 1-shot segmentation.}\label{ablation1shot}
	\end{subtable}
	\quad
	\begin{subtable}[t]{3.3in}
		\centering
		\begin{tabular}{lcccccc}
			\toprule[1pt]
			\toprule[1pt]
			+SSM & +SS Loss & i=0 & i=1 & i=2 & i=3 & mean\\
			\midrule
			& & 53.2 & 66.5 & 55.5 & 51.4 & 56.7  \\
			\checkmark & & 56.6  & 67.2 & 60.4 & 54.0 & 59.6  \\
			& \checkmark & 56.8 & 68.7  & 61.4  & 55.0 & 60.5  \\
			\checkmark & \checkmark & \textbf{58.6}  & \textbf{68.7} & \textbf{62.9} & \textbf{55.3} & \textbf{61.4} \\
			\bottomrule[1pt]
			\bottomrule[1pt]
		\end{tabular}
		\caption{Results for 5-shot segmentation.}\label{ablation5shot}
	\end{subtable}
	\caption{Ablation study on PASCAL-$5^{i}$ dataset under the metric of mean-IoU. Bold fonts represent the best results.}\label{ablationshot}
\end{table}

\subsection{Deep Learnable Meta Learner}
\label{section:meta}
Inspired by the Relation Network, we apply a deep non-linear metric to measure the similarity between the feature map of query image and the descriptor generated by base learner. As in \cite{rakelly2018few}, we also use the deep learnable metric with late fusion as the main component of meta learner. First, we multiply the features by the downsampling mask and aggregate it to obtain the latent features of the foreground. This feature is then tiled to the original spatial scale, so that each dimension of the query feature
is aligned with the representative feature ($R_{s}^{r}$):
\begin{equation}
	R_{s}^{r} = tile(pool(R_{s}^{\prime} \cdot T_{S}^{i}))
\end{equation}
Through the Relation Network comparator, the response area of the query image is obtained:
\begin{equation}
	M = Relation(R_{q}, R_{s}^{r})
\end{equation}
Finally, we feed it into the segmentation decoder, refining
and restoring the original image size to get accurate segmentation results.

\subsection{Loss and Generalization to 5-shot setting}
In addition to the cross-entropy loss function of segmentation obtained by query-support set commonly used in 
other methods (main loss), we also include the cross-entropy loss of support-support segmentation 
from base learner (auxiliary loss) into the final training loss as an auxiliary. In our method, the auxiliary 
loss itself is created as an intermediate process, so there is no much extra computation required. It can be seen 
from the experiment part that the auxiliary loss can accelerate the convergence and improve the performance.

When generalizing to 5-shot segmentation, the main difference is the base learner part. Considering that the number 
of samples for 5-shot segmentation is still small, if we calculate the gradient optimization together, it is easy 
to produce large discrepancy. Therefore, we calculate the gradient value separately, getting 5 separate 
response areas, and then take the weighted summation according to the self-supervised scores to get the final 
result, that is:
\begin{equation}
	M_{weighted} = \sum_{i=1}^{5} f_{IoU}(S_{sup}^{i}, T_{S}^{i}) \cdot f_{m}(R_{q}^{i}, R_{s}^{i \prime};\theta_{m})
\end{equation}
where $f_{IoU}$ represents the function which is used to calculate the IoU scores. 

\begin{figure}[t]
	\centering
	\includegraphics[width=78mm]{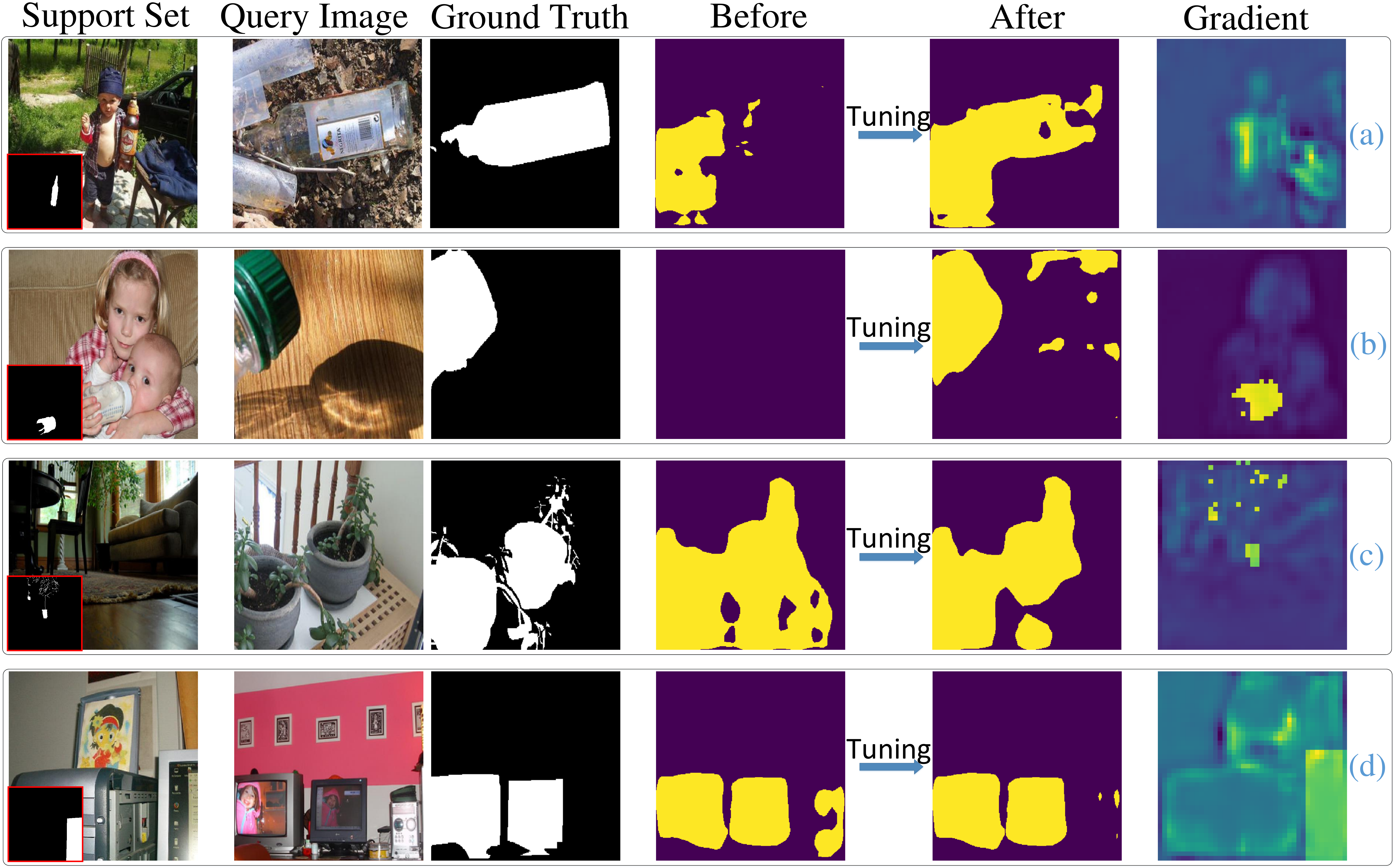}\\
	\caption{Visualization before and after applying the self-supervised module. From left to right in
	each row, they represent the support set, query image, the ground truth, two different segmentation
	results and the gradient information. The support mask is placed in the left-down corner of the support
	image.}
	\label{fig:SSM}
\end{figure}

\begin{figure}[t]
	\centering
	\includegraphics[width=78mm]{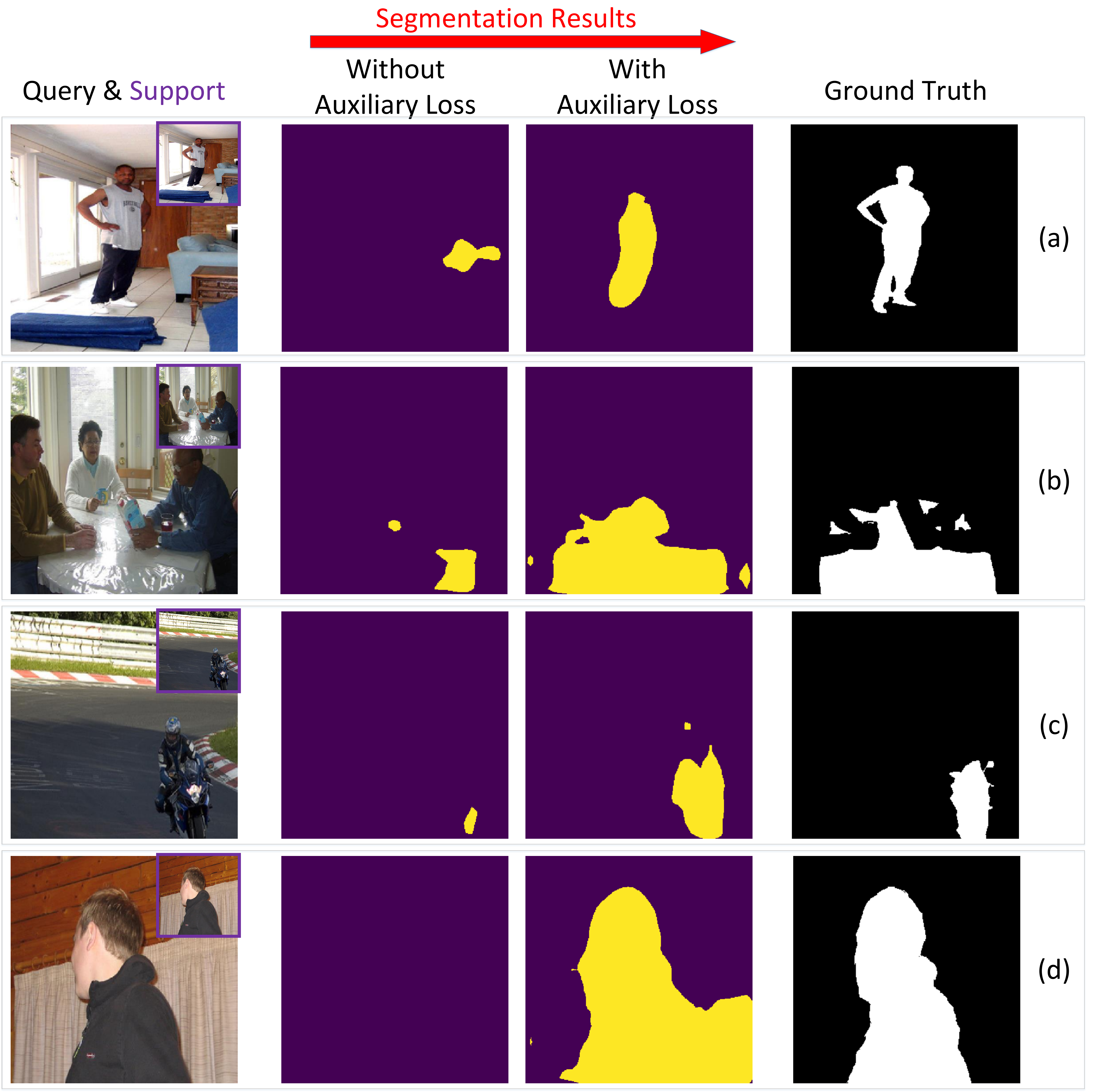}\\
	\caption{Four sets of self-supervised results from different models trained with and without the auxiliary loss. The query
	image and support image (in the upper-right corner of the first image) are the same in each row.}
	\label{visual:subloss}
\end{figure}

\begin{figure}[t]  
	\centering  
	\includegraphics[width=78mm]{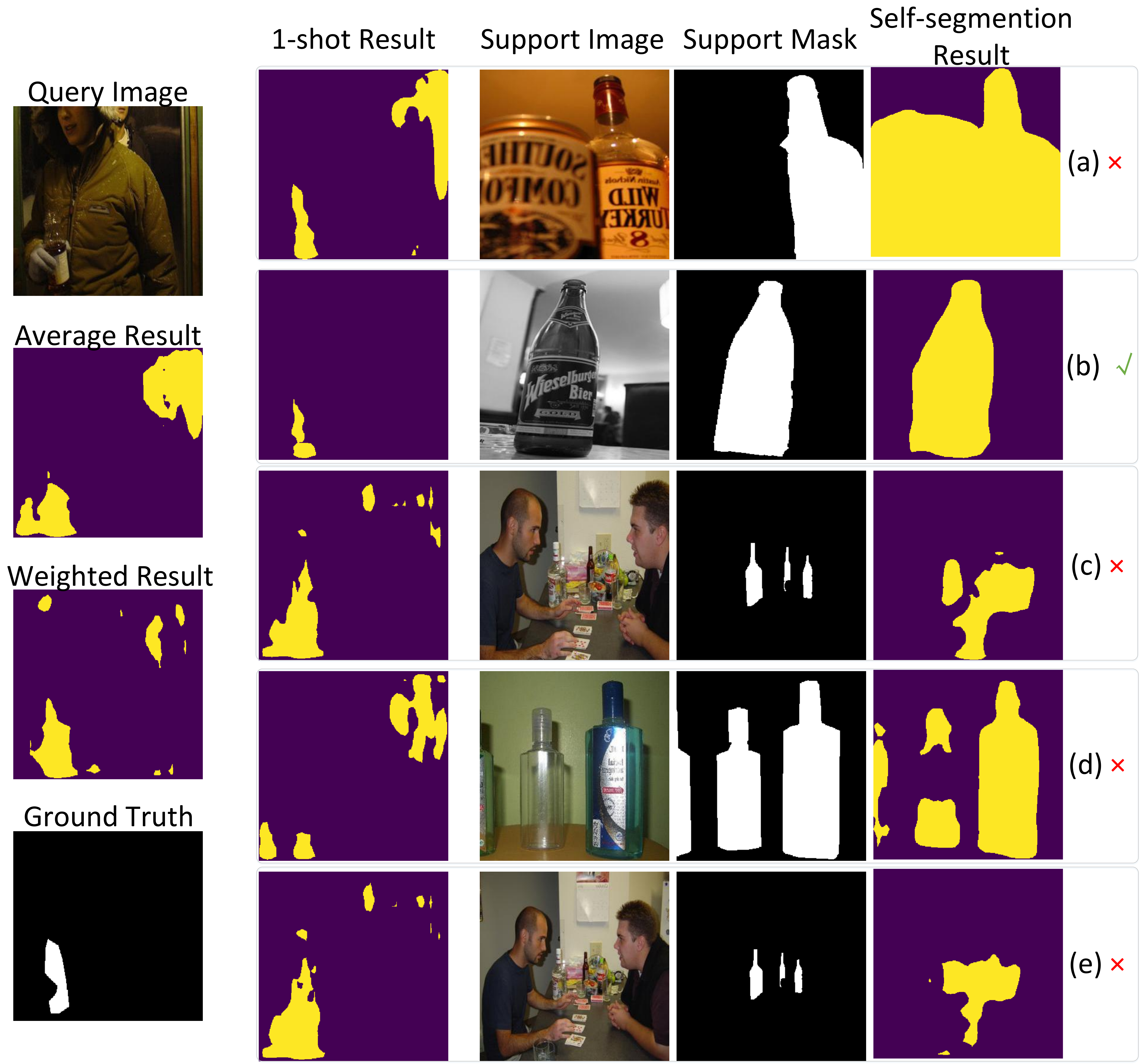}\\  
	\caption{5-shot segmentation results. The query image and the 5-shot results are placed in the first column, and the 
	1-shot results of 5 support images are in the second column. The green tick and red crosses represent right and 
	wrong segmentation results, respectively.}
	\label{visual:5shot}  
\end{figure}

\begin{table}[t]
	\centering
	\begin{tabular}{cc}
	\toprule[1pt]
	\toprule[1pt]
	Methods  & Results(mean-IoU \%)   \\ \hline
	Maximum  & 59.4          \\ 
	Average  & 61.2          \\ 
	\midrule
	Weighted & \textbf{61.4} \\
	\bottomrule[1pt]
	\bottomrule[1pt]
	\end{tabular}
	\caption{Comparison among different fusion methods in 5-shot setting under the metric of mean-IoU.}
	\label{table:fusion}
\end{table}



\section{Experiment}
\subsection{Dataset and Settings}
\textbf{Dataset.} To evaluate the performance of our model, we experiment on PASCAL-$5^{i}$ and COCO-$20^{i}$ 
datasets. The former is first 
proposed in \cite{shaban2017one} and is recognized as the standard dataset in the field of few-shot segmentation 
in subsequent work. That is, from the set of 20 classes on PASCAL dataset, we sample five and consider them as the 
test subsets $U_{test_{i}}=\left\{ 4i+1,\cdots4i+5 \right\}$, with i being the fold number ($i=0,1,2,3$), and the 
remaining 15 classes form the training set $U_{train_{i}}$. COCO-$20^{i}$ dataset is proposed in recent work and 
the division is similar to PASCAL-$5^{i}$ dataset. In the test 
stage, we randomly sample 1000 pairs of images from the corresponding test subset.

\textbf{Settings.} As adopted in \cite{shaban2017one}, we choose the per-class foreground Intersection-over-Union (IoU) 
and the average IoU over all classes (mean-IoU) as the main evaluation indicator of our task. While the foreground 
IoU and background IoU (FB-IoU) is a commonly used indicator in the field of binary segmentation, it is used by 
few papers of few-shot segmentation task. Because mean-IoU can better measure the overall performance of 
different classes and ignore the proportion of background pixels, we show the results of mean-IoU in all experiments. 

The backbone of the existing methods are different, mainly VGG-16 \cite{simonyan2014very} and ResNet-50 \cite{he2016deep}. To make a fair comparison, we 
separately train two models with different backbones for testing.

Our model uses the SGD optimizer during the training process. The initial learning rate is set to 0.0005 and the 
attenuation rate is set to 0.0005. The model stops training after 200 epochs. All images are resized to $321\times 321$ 
size and the batch size is set to 16.

\subsection{Ablation Study}

To prove the effectiveness of our architecture, We conduct several ablation experiments on PASCAL-$5^{i}$ dataset as shown 
in Table \ref{ablationshot}. To improve efficiency, we only choose ResNet50 as backbone for all models in the ablation 
study for fair comparison. The performance of our network is mainly attributed 
to two prominent components: SSM and auxiliary loss. Note that we progressively add addition components to the baseline, 
which enables us to gauge the performance improvement obtained by each of them.
Due to the fact that the self-supervised tuning mechanism dynamically adjusts the marginalized distributions of latent  
features at each stage, SSM brings about a 1.2 and 2.9 percent mean-IoU increase in 1-shot and 5-shot settings, 
respectively. At the same time, we can see that auxiliary loss boosts the overall performance, 
resulting in a 2.3 and 3.8 percent improvement.

\begin{table}[t]
	\begin{tabular}{ccccccc}
		\toprule[1pt]
		\toprule[1pt]
		Method & Backbone  & i=0  & i=1  & i=2  & i=3  & mean \\
		OSLSM  & Vgg16     & 33.6 & 55.3 & 40.9 & 33.5 & 43.9 \\
		SG-One & Vgg16     & 40.2 & 58.4 & 48.4 & 38.4 & 46.3 \\
		PAnet  & Vgg16     & 42.3 & 58.0 & 51.1 & 41.2 & 48.1 \\
		FWBFS  & Vgg16     & 47.0 & 59.6 & 52.6 & 48.3 & 51.9 \\
		Ours   & Vgg16     & \textbf{50.9} & \textbf{63.0} & \textbf{53.6} & \textbf{49.6} & \textbf{54.3} \\ \\
		\midrule
		CAnet  & ResNet50  & 52.5 & 65.9 & 51.3 & 51.9 & 55.4 \\
		FWBFS  & ResNet101 & 51.3 & 64.5 & 56.7 & 52.2 & 56.2 \\
		Ours   & ResNet50  & \textbf{54.4} & \textbf{66.4} & \textbf{57.1} & \textbf{52.5} & \textbf{57.6} \\
		\bottomrule[1pt]
		\bottomrule[1pt]
	\end{tabular}
	\caption{Comparison with SOTA for 1-shot segmentation under the mean-IoU metric on PASCAL-$5^{i}$ dataset. Bold fonts represent the best results.}
	\label{table:SOTA1shot}
\end{table}

\begin{figure}[t]	
	\centering
	\includegraphics[width=72mm]{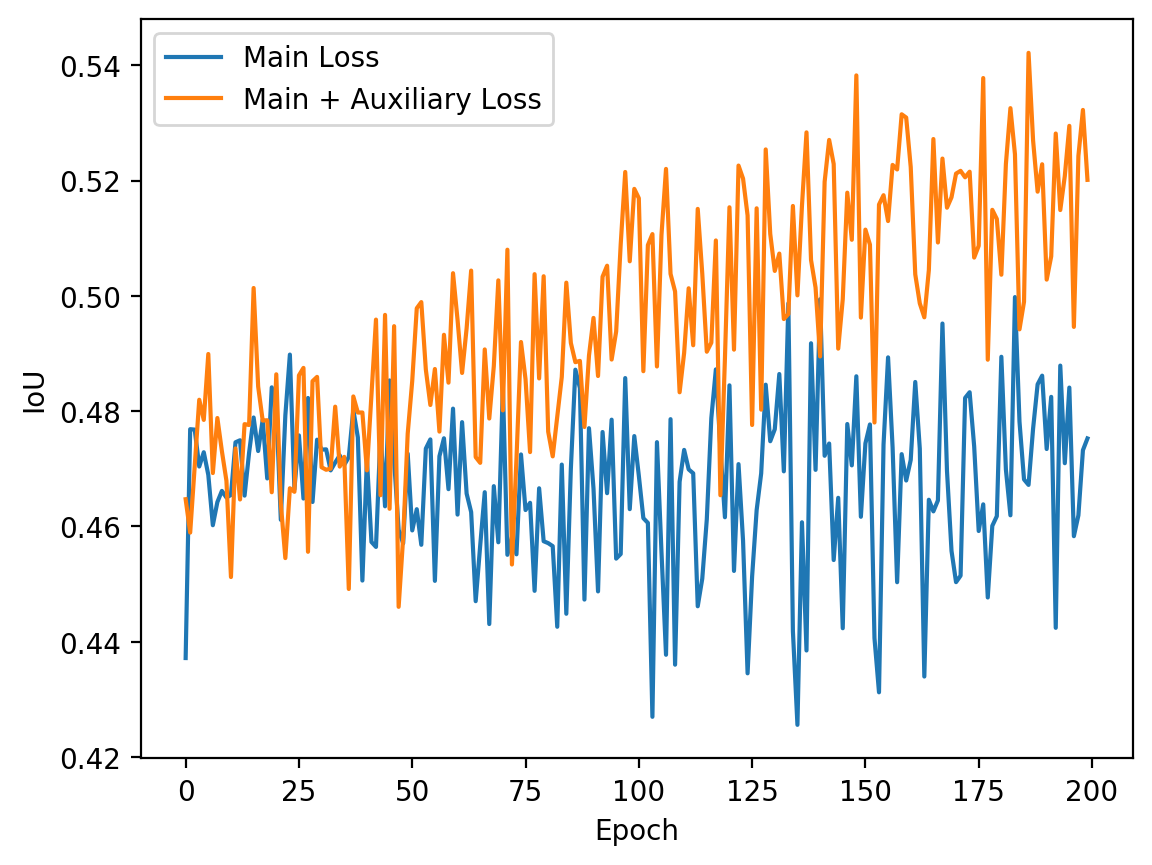}\\  
	\caption{Curves of IoU results in test stage when different loss functions are applied.}
	\label{subloss1}		
\end{figure}

\begin{table}[t]
	\begin{tabular}{ccccccc}
		\toprule[1pt]
		\toprule[1pt]
		Method & Backbone  & i=0  & i=1  & i=2  & i=3  & mean \\
		OSLSM  & Vgg16     & 35.9 & 58.1 & 42.7 & 39.1 & 43.8 \\
		SG-One & Vgg16     & 41.9 & 58.6 & 48.6 & 39.4 & 47.1 \\
		PAnet  & Vgg16     & 51.8 & 64.6 & \textbf{59.8} & 46.5 & 55.7 \\
		FWBFS  & Vgg16     & 50.9 & 62.9 & 56.5 & 50.1 & 55.1 \\
		Ours   & Vgg16     & \textbf{52.5} & \textbf{64.8} & 59.5 & \textbf{51.3} & \textbf{57.0} \\ \\
		\midrule
		CAnet  & ResNet50  & 55.5 & 67.8 & 51.9 & 53.2 & 57.1 \\
		FWBFS  & ResNet101 & 54.8 & 67.4 & 62.2 & 55.3 & 59.9 \\
		Ours   & ResNet50  & \textbf{58.6} & \textbf{68.7} & \textbf{63.1} & \textbf{55.3} & \textbf{61.4} \\
		\bottomrule[1pt]
		\bottomrule[1pt]
	\end{tabular}
	\caption{Comparison with SOTA for 5-shot segmentation under the mean-IoU metric on PASCAL-$5^{i}$ dataset. Bold fonts represent the best results.}
	\label{table:SOTA5shot}
\end{table}

\begin{table}[t]
	\centering
	\setlength{\tabcolsep}{5mm}{
	\begin{tabular}{c|cc}
	\toprule[1pt]
	\toprule[1pt]
	\multirow{2}{*}{Method} & \multicolumn{2}{c}{mean-IoU} \\ \cline{2-3} 
	& 1-shot        & 5-shot       \\ \hline
	PANet           & 20.9          & 29.7         \\
	FWBFS           & 21.2          & 23.7         \\
	Ours            & \textbf{22.2}         & \textbf{31.3}             \\ 
	\bottomrule[1pt]
	\bottomrule[1pt]
	\end{tabular}	}
	\caption{Comparison with SOTA under the mean-IoU metric on COCO-$20^{i}$ dataset.}
	\label{table:SOTAcoco}
\end{table}

\subsection{Analysis}
As PASCAL-$5^{i}$ is the most commonly used dataset by all few-shot segmentation methods, the main analysis parts of 
our experiment are accomplished on PASCAL-$5^{i}$ dataset.
\subsubsection{Effect of the Self-Supervised Module}

To clarify the actual function of the self-supervised module for few-shot segmentation task, we show the following 
visualization results. As shown in Fig. \ref{fig:SSM}, we feed the feature representations generated before and after the tuning 
process to Relation Network to obtain two sets of segmentation results. Note that the original network often 
segmentes other objects by mistakes that are easily confused in the background. After adding the SSM module, 
the category-specific semantic constraint is introduced to help query images to correct the prediction results. 
To demonstrate the meaning of this semantic constraint, the calculated gradient is visualized in the last column. We can see 
that it strengthens the focus on the regions of target categories.

\subsubsection{Importance of Auxiliary Loss}
To verify the role of auxiliary loss in the training stage, we designed the following experiment. First, we show the 
mean-IoU curves generated with and without our auxiliary loss during test stage in Fig. \ref{subloss1}. 
It can be clearly seen that the convergence is faster and the mean-IoU result is 
better after the loss is added. Then, we visualize the segmentation results of two models trained with and without 
auxiliary loss. As shown in Fig. \ref{visual:subloss}, the model without auxiliary loss can not segment 
the support images themselves, therefore inapplicable for other images of the same category. The proposed auxiliary loss
improves the self-supervised capability as well as the performance of few-shot segmentation.

\subsubsection{Comparison among different 5-shot fusion methods}
To prove the superiority of our weighted fusion in 5-shot setting, we compare the 5-shot segmentation results with 
the average and maximum fusion methods, in which the average and maximum segmentation results of 5 support images
are computed, respectively. It can be seen in Table \ref{table:fusion} that our weighted fusion strategy achieves the best. 
The samples in the PASCAL-$5^{i}$ dataset are relatively simple, and most of the weights are close to one fifth, 
so the result of average fusion method is similar to weighted fusion in general.

\subsection{Comparsion with SOTA}
To better assess the overall performance of our network, we compare it to other methods (OSLSM \cite{shaban2017one}, 
SG-One \cite{zhang2018sg}, PAnet \cite{wang2019panet}, FWBFS \cite{nguyen2019feature} and CAnet \cite{zhang2019canet}) 
on PASCAL-$5^{i}$ and COCO-$20^{i}$ datasets. 
\subsubsection{PASCAL-$5^{i}$}
We train two types of SST (Self-Supervised Tuning) models with VGG-16 and ResNet-50 backbones (we call them SST-vgg 
and SST-res model) on PASCAL-$5^{i}$ dataset. It can be seen that our SST-vgg model surpasses the best existing method over two percentage 
points in 1-shot setting, and the SST-res model yields 2.2 points improvement, which is even 1.4 points higher than 
the method with ResNet-101 backbone. Under the setting of 5-shot, our SST-vgg and SST-res models significantly 
increase by 1.9 and 1.5 points, respectively. These comparisons indicate that our method boosts the recognition 
performance of few-shot segmentation.

\subsubsection{COCO-$20^{i}$}

To prove that our method also has good generalization performance on larger datasets, we compare our method with 
others which recently report results on the COCO-$20^{i}$ dataset. Obviously, the average results of our method surpass
other best methods by 1 and 1.6 points under 1-shot and 5-shot setting, respectively.


\section{Conclusion}
In this paper, a self-supervised tuning framework is proposed for few-shot segmentation. The category-specific semantic 
constraint is provided by the self-supervised inner loop and utilized to adjust the distribution of latent features across 
different episodes. The resulting auxiliary loss is also introduced into the outer loop of training process, achieving faster 
convergence and higher scores. Extensive experiments on benchmarks show that our model is superior in both performance and 
adaptability compared with existing methods.








\bibliographystyle{named}
\bibliography{ijcai20}

\end{document}